\pdfoutput=1
\documentclass[letterpaper, 10 pt, conference]{ieeeconf}
\IEEEoverridecommandlockouts
\let\labelindent\relax

\usepackage{amsmath,amsfonts,amssymb}
\usepackage{mathtools}
\usepackage{array}

\usepackage[labelformat=simple]{subfig}
\usepackage[font={footnotesize}]{caption}
\usepackage{textcomp}
\usepackage{verbatim}
\usepackage{graphicx}
\usepackage{cite}
\usepackage{gensymb}
\usepackage{booktabs}
\usepackage{float}
\usepackage{mathrsfs}
\usepackage{bm}
\usepackage{makecell}
\usepackage{color,xcolor}
\usepackage{babel}
\usepackage{booktabs}
\usepackage{threeparttable}
\usepackage{url}
\usepackage{diagbox}
\usepackage{stfloats}
\usepackage{amsopn}
\usepackage{multirow}
\usepackage{enumitem}

\usepackage{tikz}
\usepackage{arydshln}
\usepackage{multirow}
\usepackage{bm}
\usepackage{epstopdf}
\usepackage{cite}
\usepackage{siunitx}
\usepackage{xcolor}
\usepackage[ruled,linesnumbered,boxed,noend]{algorithm2e}
\usepackage{booktabs}
\usepackage{graphicx}
\usepackage{threeparttable}
\usepackage{stfloats}

\makeatletter
\let\NAT@parse\undefined
\makeatother

\usepackage[
            citecolor=black,
            ]{hyperref}

\allowdisplaybreaks[4]

\title{\LARGE\bf
RH-Map: Online Map Construction Framework of Dynamic Object Removal Based on 3D Region-wise Hash Map Structure 
}
\vspace{-0.7cm}

\author{Zihong Yan, Xiaoyi Wu, Zhuozhu Jian, Bin Lan, and Xueqian Wang
\thanks{
This work was supported by the National Natural Science Foundation of China under Grant 62293545.
\textit{(Zihong Yan and Xiaoyi Wu are co-first authors.) (Corresponding authors: Xueqian Wang.)}}
\thanks{Zihong Yan, Zhuozhu Jian, Xueqian Wang are with the Center for Artificial Intelligence and Robotics, Shenzhen International Graduate School, Tsinghua University, Shenzhen 518055, China (e-mail: 
\href{mailto:yanzh22@mails.tsinghua.edu.cn}{yanzh22@mails.tsinghua.edu.cn};
\href{mailto:jzz21@mails.tsinghua.edu.cn}{jzz21@mails.tsinghua.edu.cn};
\href{mailto:wang.xq@sz.tsinghua.edu.cn}{wang.xq@sz.tsinghua.edu.cn};).}
\thanks{Xiaoyi Wu is with the School of Mechanical Engineering and Automation at Harbin Institute of Technology, Shenzhen 518055, China (e-mail: 
\href{mailto:200320106@stu.hit.edu.cn}{200320306@stu.hit.edu.cn})).}\thanks{Bin Lan is with Jianghuai Advance Technology Center 230022, China(e-mail: 
\href{mailto:lanbin.thu@gmail.com}{lanbin.thu@gmail.com}}
}

\begin{document}
\maketitle
\begin{abstract}
Mobile robots navigating in outdoor environments frequently encounter the issue of undesired traces left by dynamic objects and manifested as obstacles on map, impeding robots from achieving accurate localization and effective navigation. To tackle the problem, a novel map construction framework based on 3D region-wise hash map structure (\textit{RH-Map}) is proposed,
consisting of front-end \textit{scan fresher} and back-end removal modules, which realizes real-time map construction and online dynamic object removal (DOR). First, a two-layer 3D region-wise hash map structure of map management is proposed for effective online DOR. Then, in \textit{scan fresher}, region-wise ground plane estimation (R-GPE) is adopted for estimating and preserving ground information and Scan-to-Map Removal (S2M-R) is proposed to discriminate and remove dynamic regions.
Moreover, the lightweight back-end removal module maintaining keyframes is proposed for further DOR. As experimentally verified on SemanticKITTI, our proposed framework yields promising performance on online DOR of map construction compared with the state-of-the-art methods. And we also validate the proposed framework in real-world environments. The source code is released to the community\footnote{\textit{RH-Map}: \url{https://github.com/YZH-bot/RH-Map}}.

\end{abstract}

\section{INTRODUCTION }
\begin{figure}[t]
    \centering
    \includegraphics[width=8.5cm]{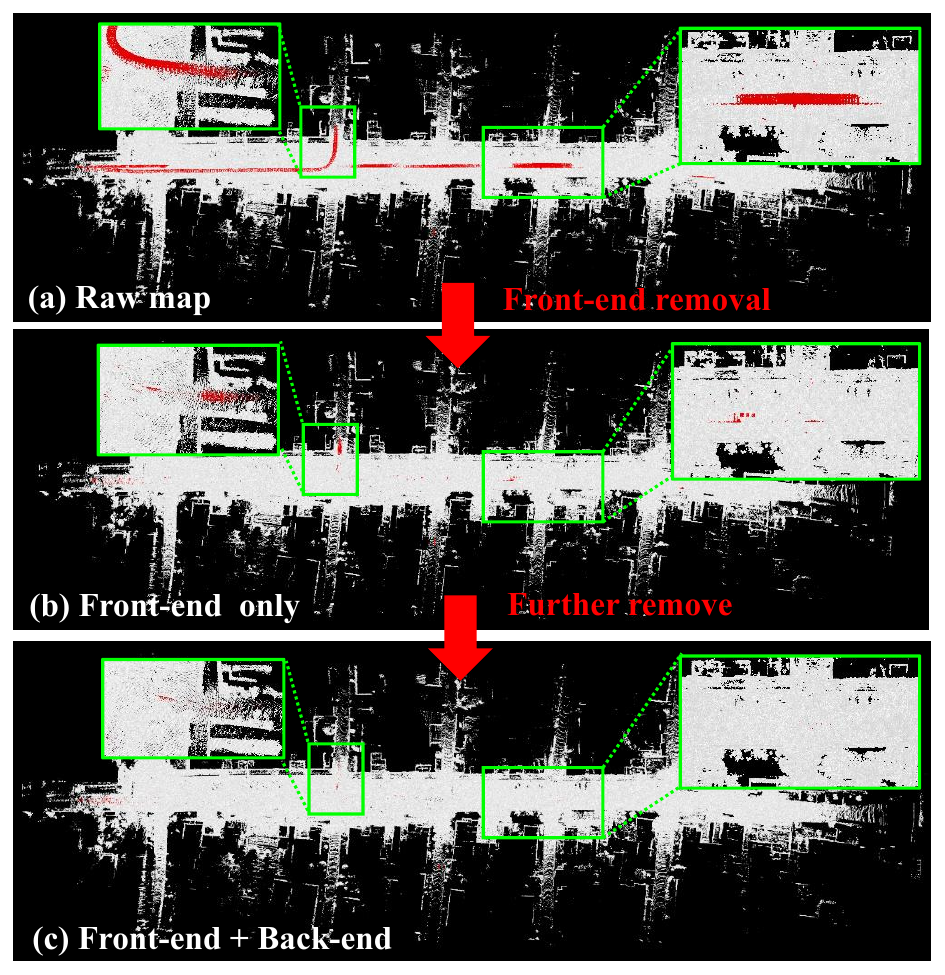}
    \caption{(a) Raw point cloud map (b) Processing via the Front-end module only (c) Further removal of dynamic objects with the Back-end module. Run at 10 Hz on sequence 05 of SemanticKITTI (dynamic objects in \textcolor{red}{red}).
}
    \label{fig_exp_sen1}
    \vspace{-0.8cm}
\end{figure}

Clean and reliable maps play an important role in navigation and exploration of mobile robot platforms such as Unmanned Ground Vehicles (UGVs) and Unmanned Aerial Vehicles (UAVs) in outdoor environments. And Simultaneously Localization and Mapping (SLAM) techniques are usually applied to provide robust localization and mapping of surroundings\cite{shan2020lio}\cite{xu2022fast}.
However, most SLAM methods assume that environments are relatively static; therefore, dynamic objects in the surroundings may leave residual traces on map as shown in Fig.\ref{fig_exp_sen1}(a), known as \textit{ghost trail effect}\cite{pagad2020robust}. 
These usually lead to confusion in environment representation and sub-optimal results of path planning. There are SLAM researches aiming at tackling this assumption by either employing online dynamic object removal (DOR)\cite{chen2019suma++,qian2022rf,park2022nonparametric} or utilizing states of dynamic objects in optimization\cite{tian2022dl} to ensure more reliable localization in dynamic environments. Nevertheless, they focus more on improving the accuracy of localization rather than providing a dense and clean map. 

In this work, we focus on real-time DOR for online dense map construction in large-scale environments using 3D LiDAR. 
Some researches tackling DOR mission in map construction typically require long sequences of historical information, submap or pre-built global map\cite{kim2020remove,lim2021erasor,fan2022dynamicfilter}, which are unsuitable for scenarios with high real-time requirements. Additionally, online DOR methods rely on short-term neighboring frames of LiDAR information, which can filter out most of the dynamic objects in the current scan. However, due to the sparsity and incidence angle ambiguity characteristics of LiDAR in distant areas, some dynamic objects may persist and remain on the map indefinitely while some static points are misclassified, thereby degrading map quality.
We summarize the challenges as follows: 1) the difficulty in maintaining the timeliness of map construction while simultaneously filtering out dynamic objects, and 2) the weak performance of online DOR due to characteristics of LiDAR data, negatively impacting the quality of the map.

This work proposes a novel online map construction framework of DOR in dynamic environments referred to as \textit{RH-Map}. First, we propose a two-layer map structure, managing map with 3D regions and minimum resolution cubes, which leverages the effectiveness of region-wise structure and $O$(1) complexity of hash map for online DOR and real-time map update. Based on the region-wise structure, a two-stage online mapping framework for constructing static map including front-end and back-end modules is introduced. The front-end module estimates ground information for preservation and removes potential dynamic regions on map utilizing the current scan. And the back-end module maintains a keyframe queue of historical information to further remove residual dynamic objects effectively as shown in Fig.\ref{fig_exp_sen1}. We summarize the contributions as follows:
\begin{enumerate}
\item A region-wise hash map structure for efficient map management is proposed, which facilitates online DOR and ensures real-time map updates, achieving real-time performance over 10Hz on SemanticKITTI.
\item RH-Map, a novel online map construction framework of DOR consisting of real-time front-end \textit{scan fresher} and lightweight back-end removal modules is proposed. \textit{Scan fresher} achieves robust ground estimation and real-time online DOR, and the back-end module realizes further removal of dynamic objects to enhance map quality.
\item We validate the proposed method on SemanticKITTI thoroughly and demonstrate the online performance in real-world scenes. Our method achieves state-of-the-art performance in DOR and can apply to robot navigation in real-world.
\end{enumerate}

\section{ RELATED WORKS}
In this section, we review the related works about DOR of static map construction, which mainly fall into three categories: occupancy map-based methods, visibility-based methods and segmentation-based methods.

\textbf{Occupancy map-based methods}:  Based on the principle of ray propagation, occupancy map-based methods update probability of ray endpoint as occupied and space traveled by ray as free by using ray-tracing\cite{hornung2013octomap}\cite{schauer2018peopleremover}. However, ray-tracing suffers from massive calculations\cite{kim2020remove}\cite{fan2022dynamicfilter}\cite{yoon2019mapless} and severe misclassification caused by incidence angle ambiguity\cite{lim2021erasor}.
Schauer and Nüchter\cite{schauer2018peopleremover} propose to traverse voxel grid to find differences in the occupancy between the scans thus estimating the voxel as dynamic or static, which alleviates the loss of ground information to some extent. Pagad \textit{et al.}\cite{pagad2020robust} performs combination of object detection and improved Octomap with hardware GPU acceleration. Despite the aforementioned algorithmic and engineering optimizations, occupancy map-based methods using ray-tracing are still challenging for real-time online DOR.

\textbf{Visibility-based methods}:
To address computational complexity of ray-tracing, visibility-based approaches have been introduced based on range image disparity between LiDAR and map to determine and remove dynamic points on map\cite{kim2020remove}. However, visibility-based still suffer from incidence angle ambiguity. Kim and Kim \cite{kim2020remove} propose a visibility-based approach based on multi-resolution range images to construct static map offline. And Fu \textit{et al.}\cite{fu2022mapcleaner} propose an inverse way of projection, i.e., they back project the map onto each frame and perform a map-to-frame comparison instead to avoid quantization error and achieved state-of-the-art performance with terrain modeling.
Compared to offline methods utilizing long sequences of information, online DOR paradigm requires real-time performance and can only rely on short sequences of real-time information, which limits the removal performance\cite{fan2022dynamicfilter}. And the sparsity issue also impacts the performance of online DOR. Therefore, Fan \textit{et al.}\cite{fan2022dynamicfilter} proposed an online DOR framework integrating submap-based and visibility-based methods and first proposed front-end and back-end framework utilizing historical information for further removal. Nonetheless, they only rely on certain thresholds to restrict the incidence angle, which cannot effectively preserve the ground.

\textbf{ Segmentation-based methods}: We further divide this category into scan pre-segmentation methods and ground segmentation methods based on the segmentation stage.

\textbf{\textit{a) Scan pre-segmentation methods}} primarily rely on frame-to-frame discrepancies or deep learning techniques to detect dynamic objects in the current scan. Yoon\textit{et al}.\cite{yoon2019mapless} utilize inter-frame Lidar information discrepancies and clustering to detect and remove dynamic regions. In addition, deep learning methods have shown great potential in providing semantic information from pre-trained neural network models and performing filtering for DOR in recent years\cite{cortinhal2020salsanext,li2021multi,chen2022automatic}. To address the issue of mis-segmenting static semantic objects in the environment, for example, parked cars, Sun et al.\cite{chen2021moving}\cite{sun2022efficient} use LiDAR semantic segmentation networks based on residual range images of sequential data to differentiate between dynamic and static objects in LiDAR scans. However, pre-segmentation methods do not focus on the establishment of static maps\cite{han2023gardenmap}, resulting in incorrectly segmented points being permanently retained on the static map, and deep learning methods heavily depend on meticulously annotated datasets.

\textbf{\textit{b) Ground segmentation methods}} refer to the process of separating ground information from the LiDAR scans\cite{dewan2016motion,asvadi20163d,lim2021patchwork,lee2022patchwork++} or maps\cite{lim2021erasor}\cite{fu2022mapcleaner}\cite{wang2022dynamic}, enabling further dynamic object detection and tracking, and preserving the integrity of ground information. 
Arora et al.\cite{arora2023static} integrate offline ground segmentation into OctoMap to improve the distinction between moving objects and static road backgrounds. And Lim et al.\cite{lim2021erasor} apply Region-wise Ground Plane Fitting (R-GPF) with low computational load for static map construction, leveraging advantages of region-wise data structure to approximating the plane. And wang et al.\cite{wang2023drr} propose a hierarchical vertical height descriptor in local perspective based on distribution description to better reflect data distribution in each region. However, both of these partitioning methods represent 3D information using 2.5D elevation descriptors, which leads to a loss of information as reflected in Fig.\ref{fig_FP}(b).

\section{IMPLEMENTATION}
\begin{figure*}[t]
    \center
    \includegraphics[width=17.5cm]{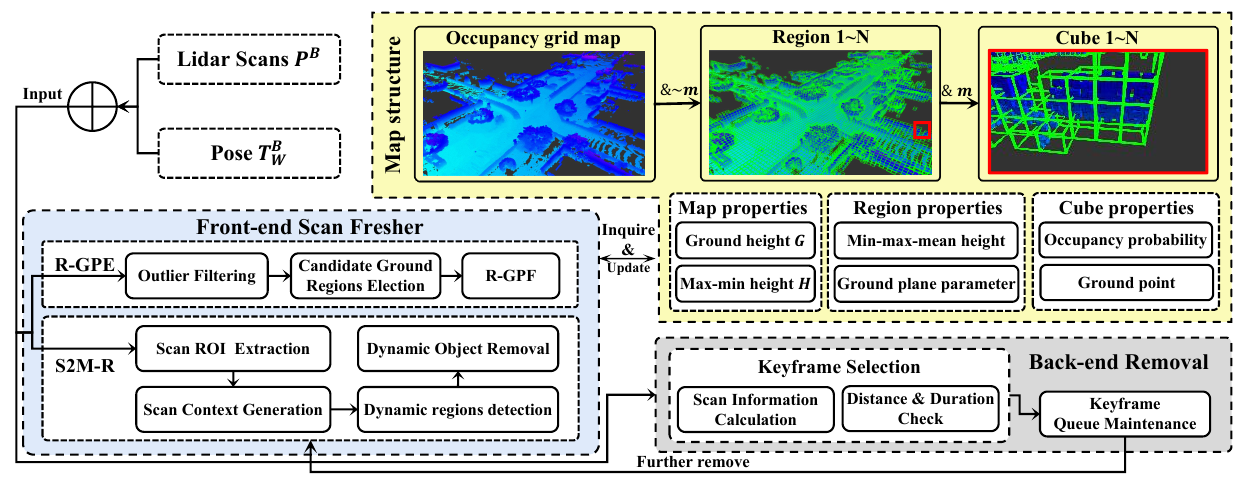}
       \vspace{-0.2cm}
    \caption{
    Overview of RH-Map: the map is constructed based on region-wise map structure as shown in the top right corner of Fig.\ref{fig_framework}. Given the current scan $P_B$ and Lidar-to-World transformation $T^W_B$ estimated from LiDAR Odometry, the Region-wise Ground Plane Estimation (R-GPE) estimates the ground plane of each region and updates the properties of map, regions and cubes. Meanwhile, Scan-to-Map Removal(S2M-R) module detects candidate regions with dynamic objects and removes dynamic objects based on the properties. With keyframe selection, the Back-end Removal module maintains keyframe queue and further removes dynamic object residues on the map.
}
    \label{fig_framework}
    \vspace{-0.5cm}
\end{figure*}

The schematic diagram of RH-Map is illustrated in Fig.\ref{fig_framework}. Based on the proposed two-layer hash map structure, RH-Map utilizes 3D regions to represent the information of local areas for DOR and each region stores state of cubes (minimum resolution) in the space. And RH-Map consists of two parts: the front-end \textit{scan fresher} and the back-end removal modules based on the proposed data structure.

\subsection{Problem Definition}
Different from previous works\cite{kim2020remove}\cite{lim2021erasor}\cite{schauer2018peopleremover}, we focus on online DOR of map constructed by consecutive LiDAR scans.
In this work, we use state-of-the-art LiDAR-based Odometry or SLAM methods such as\cite{shan2020lio}\cite{xu2022fast} to provide SE(3) pose $T^W_B$ between world frame $W$ and LiDAR frame $B$, and receive real-time scan $P^B$ as input.

\subsection{Region-wise hash map structure}
\label{subsec: Region-wise map structure}

\begin{figure}[t]
    \centering
    \includegraphics[width=8.5cm]{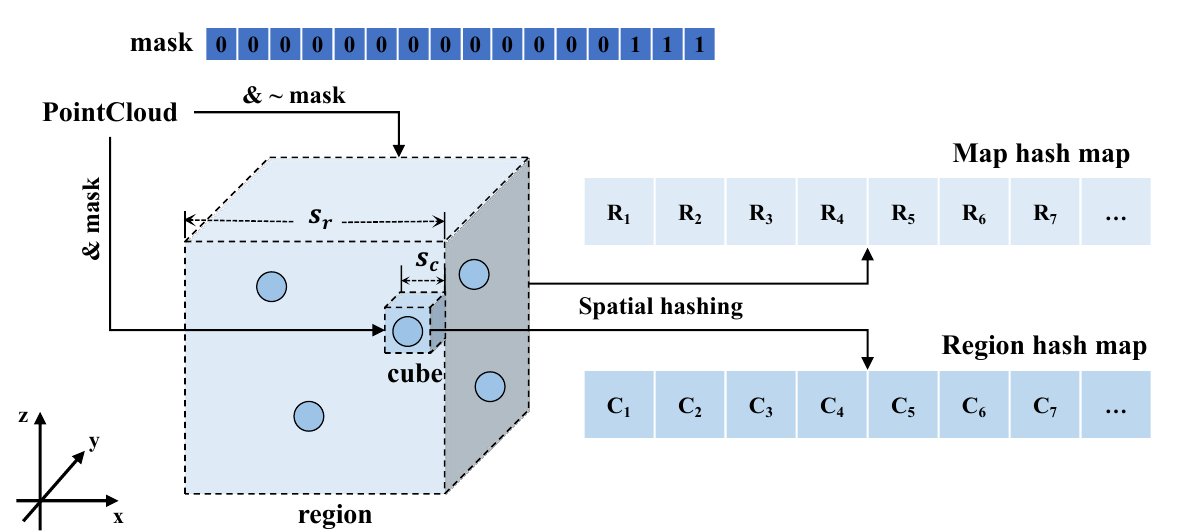}
    \caption{The structure of the two-layer region-wise hash map.}
    \label{map_structure}
    \vspace{-0.5cm}
\end{figure}

In this section, we introduce a two-layer hash map structure as shown in Fig. \ref{fig_framework}, unlike ERASOR's 2D egocentric Region-wise Pseudo Occupancy Descriptor (R-POD)\cite{lim2021erasor}, the global 3D occupancy grid map is divided into regions with uniform size (green box, 0.8m in our experiment), and each region is further divided into minimum resolution occupancy grids (Cubes, 0.1m).
In the region-wise map structure (see Fig.\ref{map_structure}), the value of binary mask $m$ and minimum resolution cube size $s_c$ determines the region size $s_r=(m+1) * s_c$ (i.e. $m=00000111, s_c=0.1m$, then $s_r=0.8m$). 
For map construction, point cloud from scan is first transformed to the world frame and converted to global index $I$, then conducted binary operation with the binary mask $m$ on each dimension to get the corresponding 3D region index $I_r$ and 3D cube index $I_c$ which compose a unique index in RH-Map:  
\begin{equation}
\begin{aligned}
\label{formula (1)}
I=\lfloor \frac{1}{s_c}[p_x, p_y, p_z]^T \rfloor,\ I_r = I\ \& \ {\raise.17ex\hbox{$\scriptstyle\sim$}} 
m, \ I_c = I\ \& \ m,
\end{aligned}
\end{equation}
where $[p_x, p_y, p_z] \in \mathbb{R}^3$ are the global coordinates of point cloud, and obviously, $I=I_r \| I_c$. Note that a 16-bit mask (in our experiment) can handle a map size of $6553.5m^3$, and expanding the mask's bit size is sufficient to accommodate a larger range. After getting the $I_r$ and $I_c$, we hash the indexes into the map using a 3D hash function like \cite{teschner2003optimized} and update the occupancy probability $p$ as follows:
\begin{equation}
\begin{split}
k_i &=(I_i(x)n_x)\ xor\ (I_i(y)n_y)\ xor \ (I_i(z)n_z)\bmod N\\
    &=hash(I_i),\ i=r,c
\end{split}
\end{equation}
\begin{equation}
\begin{split}
M(k_r,k_c)_t.p=M(k_r,k_c)_{t-1}.p+L(measurements)
\end{split}
\end{equation}
 where $I(\cdot)$ represents obtaining the value of $x,y$  or $z$ dimension of index $I$, $n_x,n_y,n_z$ are large prime numbers and $N$ is the size of our map, respectively, $L(\cdot)$ represents log odds. 

For simplification, we use $M_I=M_{I_r,I_c}$ to represent the unique cube on map with index $I=I_r\|I_c$  and $M_{I_r}$ to represent the $I_r$ region on map. And we use $O=(I(x),I(y)),O_r=(I_r(x),I_r(y))$ to denote 2D index of $I,I_r$ on $xy$ plane.
RH-Map records the highest and lowest points on 2D region index $O_r$ in $H\{max,min\}$ and the average height of ground points on 2D global index $O$ in $G$, as well as the min-max-mean height and plane parameters of each region. It also updates the occupancy probability and ground attribute of cubes. 
Our method supports online operation and the hash map structure only updates the state of the occupied places and is free from a presetting range of maps like traditional occupancy grid map or octomap\cite{hornung2013octomap}, while R-POD\cite{lim2021erasor} can only be applied in an offline manner due to the irregularity of data organization, and the time consumed will increase as the global map becomes larger.

\subsection{Scan fresher front-end}
The front-end \textit{scan fresher} contains \textit{Region-wise Ground Plane Estimation} (R-GPE) and \textit{Scan-to-Map Removal} (S2M-R). Our \textit{scan fresher} realizes online DOR in S2M-R and applies R-GPE to preserve the integrity of the ground based on our region-wise map structure mentioned in \ref{subsec: Region-wise map structure}.

\subsubsection{Region-wise Ground Plane Estimation}
\label{subsec: PGPF}

\begin{figure*}[ht]
    \center
    \includegraphics[width=17.5cm]{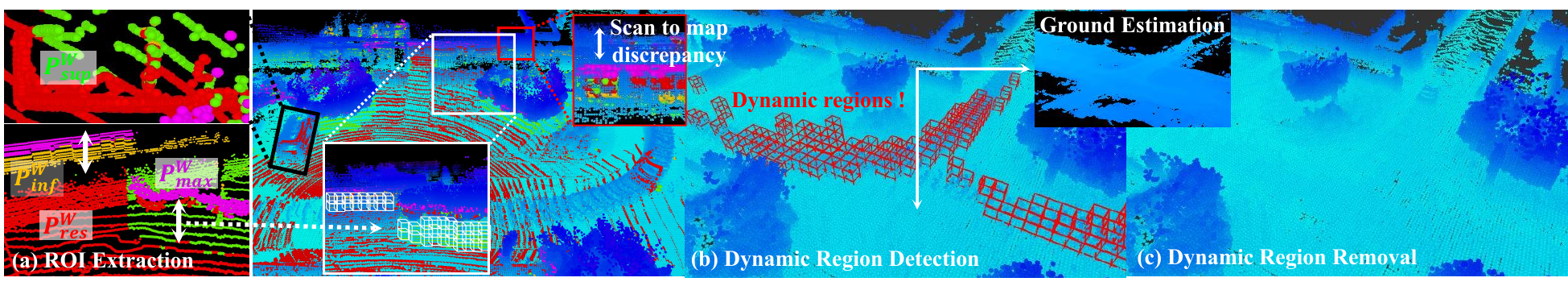}
    \caption{Procedure of S2M-R. (a) illustrates the Regions of Interest (ROI) extraction process: the max ring points \textcolor{magenta}{$P^W_{max}$} denote the maximum FOV of LiDAR, and the points \textcolor{orange}{$P^W_{inf}$} and \textcolor{green}{$P^W_{sup}$} indicate the points that use 3D \textit{scan context} for detection (b) displays the result of R-GPE and the dynamic regions in \textcolor{red}{red} indicating the presence of dynamic objects (c) demonstrates the results after S2M-R.
}
    \label{S2M-R}
    \vspace{-0.3cm}
\end{figure*}


Firstly,  real-time scan data $P^B$ is transformed into world frame $P^W=T^W_B P^B$ and inserted into our map to update $H$ of map and max-min-mean height properties of regions for detection in Section\ref{subsec:S2M-R}.
Next, we conduct \textit{candidate ground regions election} to determine the range within which ground plane fitting needs to be performed. Moreover, there are many noise points in scans that have a great influence on ground estimation. Therefore, we introduce an \textit{outlier filtering} process considering the region as noise if there is no ground within its eight adjacent regions. This is established on the assumption that ground is typically continuous and noise points usually appear discretely. Then, the candidate ground regions set $^{cgr}\mathbb{I}_r$ is obtained as follows: 
\begin{equation}
\begin{aligned}
    ^{cgr}\mathbb{I}_r= & \{ I_r^i \ | \ {I}_r^i=I^i\ \| \ m,I^i\in \mathbb{I}^{P^W},
    \\ & I^i(z) < G(O^i), M^N_{I_r^i}\neq \varnothing\ \}
\end{aligned}
\end{equation}
where $\mathbb{I}^{P^W}$ represents the set of global indexes converted from $P^W$ by formula.\ref{formula (1)}, $M^N_{I_r^i}$ denotes eight adjacent regions of $M_{I_r^i}$, and $O^i=(I^i(x),I^i(y))$.

Then, similar with the region-wise ground plane fitting (R-GPF) in \cite{lim2021erasor}, we perform R-GPF on each region $M_{I^i_r}$ ($I^i_r \in {^{cgr}\mathbb{I}_r}$). For occupied cubes smaller than mean height of $M_{I^i_r}$ in z-axis, we extract their global indexes as initial ground points set  $\mathbb{I}^{I^i_r}$. Then we employ principal component analysis (PCA) to fit the ground plane and update ground plane parameters of region $M_{I^i_r}$ with normal vector $n_{i}$ and plane coefficient $d_{i}=n_{i}\overline{I^i}$ where $\overline{I^i}$ denotes the average index of $\mathbb{I}^{I^i_r}$. And ground global indexes set $^{gro}\mathbb{I}^{I^i_r}$ of region $M_{I^i_r}$ is extracted as follows:
\begin{equation}
^{gro}\mathbb{I}^{I^i_r}=\{I^j \ | \ I^j \in \mathbb{I}^{I^i_r}, |n_{i}I^j-d_{i}|<r_{gro}\}
\end{equation}
where $r_{gro}$ denotes the distance margin of the plane. Finally, we update the ground property of cubes with $^{gro}\mathbb{I}^{I^i_r}$. Note that R-GPE will be performed only if new candidate ground regions are detected and the effect of R-GPE is demonstrated in Fig.\ref{S2M-R}(b).

\subsubsection{Scan-to-Map Removal}
\label{subsec:S2M-R}
\begin{figure}[t]
    \centering
    \includegraphics[width=8.5cm]{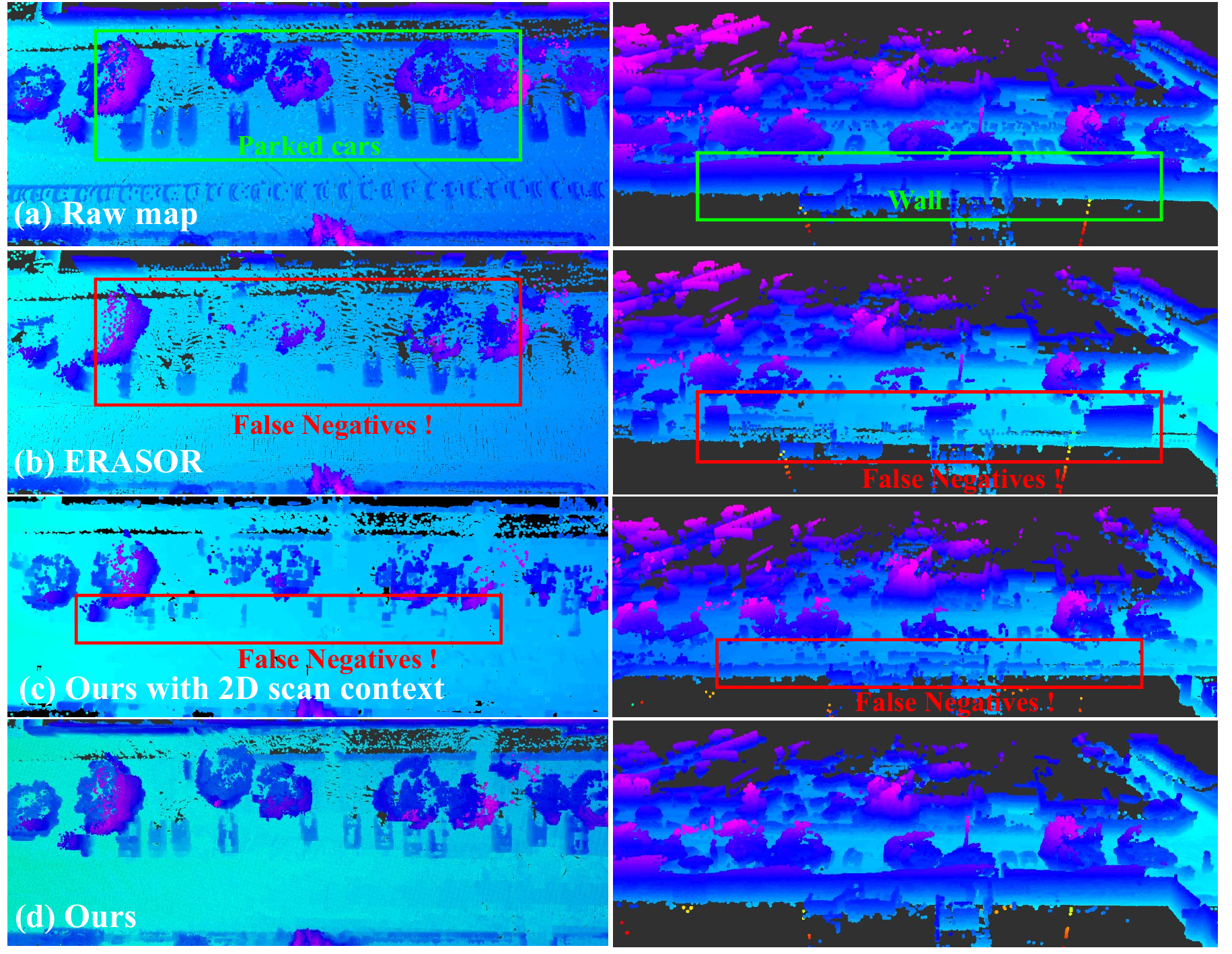}
    \caption{Effect of ROI extraction on sequence 05 of SemanticKITTI. (a) Raw map (b) ERASOR\cite{lim2021erasor}  (c) Ours with 2D scan context (d) Ours with 2D\&3D scan contexts.}
    \label{fig_FP}
    \vspace{-2.0em}
\end{figure}

Generally, information from a single scan is relatively sparse and information-deprived compared to map. And without pre-processing of LiDAR data in advance, region-wise removal methods usually lead to instances of false positives (FPs, falsely classify static points as dynamic points) such as parked cars, walls or trees as shown in Fig.\ref{fig_FP}(b) in red rectangle. And 2D descriptors like ERASOR's R-POD\cite{lim2021erasor} may result in false negatives (FNs, fail to detect dynamic points) such as dynamic objects located between the highest and lowest points. To solve these problems, we introduce Regions of Interest (ROI) extraction of $P^B$ and 2D and 3D \textit{scan context} for detecting dynamic objects of different extracted points. 

First of all, the FOV of LiDAR is relatively limited in the vicinity of LiDAR compared to the map, causing a discrepancy between LiDAR observation and map (see red rectangle in Fig.\ref{S2M-R}(a)), so the max ring of LiDAR points $P^B_{max}$ is extracted from $P^B$ to represent the maximum FOV of LiDAR: 
\begin{equation}
    P^B_{roi}=\{P^B_{res},P^B_{max}\}
\end{equation}
where $P^B_{res}$ (in red) represents the rest points of $P^B$.
And then scan $P^B_{roi}$ is then transformed to world frame $P^W_{roi}=T^W_BP^B_{roi}=\{P^W_{res},P^W_{max}\}$. 

To detect the differences between scan and map for dynamic objects detection, we introduce 2D region-wise \textit{scan context} $^{2d}S^W$ defined as followed:
\begin{equation}
^{2d}S^W_{O_r^i}=\{ Z_{max},\ Z_{min}\},O_r^i \in \mathbb{O}_r^{P^W_{roi}}
\end{equation}
where $\mathbb{O}_r^{P^W_{roi}}$ represents the set of 2D region indexes of $P^W_{roi}$, $Z_{max}$ and $Z_{min}$ denote the maximum and minimum height at 2D region index $O_r^i$, respectively. And we calculate the $ratio_1$ between  $^{2d}S^W$ and the corresponding area of map $M$ as followed:
\begin{equation}
ratio_1^{O_r^i}=
\begin{cases}
\frac{\Delta H}{H(O_r^i).max-H(O_r^i).min}\\
\frac{\Delta H}{{^{2d}S}^W_{O_r^i}.Z_{max}-H(O_r^i).min}, \exists \ {O_r}^{p_k}==O_r^i
\end{cases}
\end{equation}
where $\Delta H={^{2d}S}^W_{O_r^i}.Z_{max}-{^{2d}S}^W_{O_r^i}.Z_{min}$, $H(O_r^i).max$ and $H(O_r^i).min$ represent the max and min height map $M$ at 2D region index $O_r^i$, and ${O_r}^{p_k}$ denotes the 2D region index of points $p_k$ ($p_k \in P^W_{max}$).
If $ratio_1^{O_r^i}$ is less than removal thresholds $\delta_1$, we consider the regions with 2D region index $ O_r^i$ to be dynamic in the FOV of LiDAR. 

As mentioned the disadvantages of 2D descriptor, we still cannot completely eliminate FPs and FNs in this way as shown in Fig.\ref{fig_FP}(c). This is because 2D descriptors cannot accurately describe the dynamic and static state of “bulge-dented” points as shown in Fig.\ref{S2M-R}(a) in green and orange.
Therefore, we further distinguish these “bulge-dented” points from $P^B_{roi}$ as $P^B_{sup}$ and $P^B_{inf}$ and introduce 3D region-wise \textit{scan context} $^{3d}S^W$ for dynamic regions detection instead.
Due to the unordered nature of point cloud data, we convert $P^B_{roi}$ into range image $\mathcal{I}^B_{m\times n}$, and $P^B_{sup}$, $P^B_{inf}$ can be obtained and record the upper bound $sup$ and lower bound $inf$ for dynamic regions detection as followed:
\begin{equation}
\begin{aligned}
\label{formula (7)}
P^B_{sup} = \{ p_k.sup=Z_{i+t} \ | \ p_k\in{P^B_{roi}},\\ 
\mathcal{I}^B(i,j)\ -\ \mathcal{I}^B(i+t,j)>r_{1} \}
\end{aligned}
\end{equation}
\begin{equation}
\begin{aligned}
\label{formula (8)}
P^B_{inf} = \{ p_k.inf=Z_{i-t} \ | \ p_k\in{P^B_{roi}},\\
\mathcal{I}^B(i,j)\ -\ \mathcal{I}^B(i-t,j)>r_{2} \}
\end{aligned}
\end{equation}
where $\mathcal{I}^B(i,j)$ denotes the range distance of $p_k$, $\mathcal{I}^B(i+t,j)$, $\mathcal{I}^B(i-t,j)$ denote searching upwards and downwards until the difference in range distance exceeds threshold $r_{1}$ and $r_{2}$, respectively, and 
$Z_{i+t}$, $Z_{i-t}$ represent the height of the point corresponding to $\mathcal{I}^B(i+t,j)$ and $\mathcal{I}^B(i-t,j)$ , $t\in[0,1...]$. And $P^B_{sup},P^B_{inf}$ is then transformed to world frame $P^W_{sup},P^W_{inf}$. Noted that $sup$ and $inf$ are also transformed to world frame.
And we define  $^{3d}S^W$ as follows:
\begin{equation}
    ^{3d}S^W_{I_r^i}=\{Z_{max},Z_{min},sup,inf\},I_r^i \in \mathbb{I}_r^{P^W_{sup}} \cup \mathbb{I}_r^{P^W_{inf}}
\end{equation}
where $\mathbb{I}_r^{P^W_{sup}},\mathbb{I}_r^{P^W_{inf}}$ represent the set of 3D region indexes of $P^W_{sup},P^W_{inf}$, $Z_{max},\ Z_{min}$ and $ sup,\ inf$ are parameters that denote the maximum, minimum height and minimum upper bound of $P^W_{sup}$, maximum lower bound of $P^W_{inf}$ within region $I_r^i$, respectively. Note that $sup$ and $inf$ are initialized with $Z_{max}$ and $Z_{min}$.
Then we iterate over regions in z-axis direction between ${^{3d}S^W_{I_r^i}}.sup$ and ${^{3d}S^W_{I_r^i}}.inf$ to select candidate regions set $\mathbb{I}^M$ (see white boxes in Fig.\ref{S2M-R}(a)) for dynamic regions detection. We calculate the $ratio_2$ between  $S^W$ and the corresponding region of map $M$ as followed:
\begin{equation}
ratio_2^{I_r^i}=
\begin{cases}
\frac{^{3d}S^W_{I_r^i}.{Z_{max}-^{3d}S^W_{I_r^i}.{Z_{min}}}}{M_{I_r^i}.{Z_{max}-M_{I_r^i}.{Z_{min}}}}, I_r^i\in \mathbb{I}_r^{P^W_{sup}}\cup\mathbb{I}_r^{P^W_{inf}}\\
0, I_r^i\in (\mathbb{I}_r^{P^W_{sup}}\cup\mathbb{I}_r^{P^W_{inf}})^\complement\cap \mathbb{I}^M
\end{cases}
\end{equation}
Obviously, $ratio_1$ and $ratio_2$ are close to 1 if the scan perfectly matches the global map. If $ratio_1^{O_r^i}<\delta_1$ and $ratio_2^{I_r^i}<\delta_2$, we consider the regions located at $ O_r^i$ and the region $M_{I_r^i}$ to be dynamic as shown in Fig.\ref{S2M-R}(b). After dynamic object detection, dynamic regions are removed excluding ground points extracted in \ref{subsec: PGPF}. The effect of combining the 2D and 3D scan contexts and the procedure of \textit{S2M-R} are demonstrated in Fig.\ref{fig_FP}(d) and Fig.\ref{S2M-R}.

\subsection{Back-end Removal}
\label{Back-end Removal}
\begin{figure}[t]
    \centering
    \includegraphics[width=8.5cm]{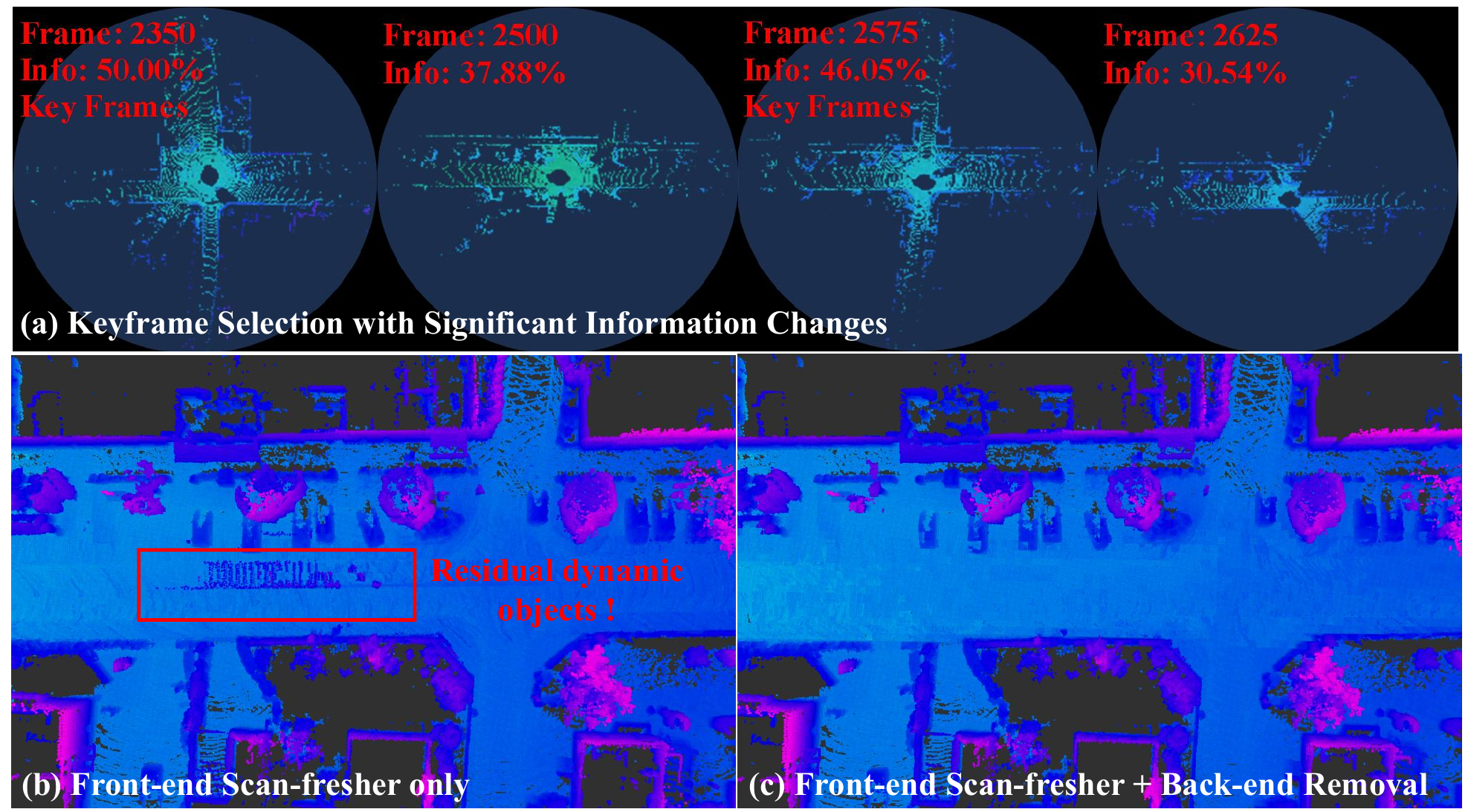}
    \caption{Back-end Removal module. (a) Information content of different LiDAR frames (b) Front-end module only with progressively accumulate residues on the map (\textcolor{red}{red} box) (c) Effect after incorporation of the Back-end Removal module component.}
    \label{fig_Back-end}
    \vspace{-0.6cm}
\end{figure}        
To obtain higher-quality static maps, historical information of sub-maps is often used for further removal because of the insufficiency of information from short-term scans. 
Regardless of methods employed, the information becomes sparser as the distance from ego center increases, ultimately resulting in inevitable presence of dynamic object residue on the map  by using short-term scans as shown in Fig.\ref{fig_Back-end}(b). Therefore, it is essential to introduce back-end to reuse historical information which is first proposed in \cite{fan2022dynamicfilter}. However, the use of sub-maps introduces a significant computational cost. In this section, we design a lightweight back-end removal module that can run at high frequency with the front-end leveraging historical information to handle residual dynamic objects.

Our back-end removal module maintains a keyframe queue, from which keyframe information is processed using the S2M-R algorithm for further DOR when the robot moves a certain distance away. The selection of keyframe primarily depends on distance, time interval and the amount of information captured by the scan. 
We define information content of a LiDAR frame as follows: 
\begin{equation}
V(\mathcal{I}_{m\times n}^B) = \frac{\sum_{j=1}^n\mathop{\max}\limits_{1\le i\le m}\ \mathcal{I}_{m\times n}^B(i,j)}{r_{max}\times n} \times 100\%
\end{equation}
where $r_{max}$ denotes the max range distance of LiDAR.
Each scan is projected into 2D plane and calculate the proportion of the blue circle occupied to represent the information content of the frame as shown in Fig.\ref{fig_Back-end}(a). The information content at an intersection is often greater than that on a single-lane road. When the information content of a frame changes significantly compared to the previous frames, it is selected as a keyframe to avoid information loss. And when the distance or time interval between current frame and previous keyframe exceeds given thresholds, it is also selected as a keyframe. The effect after the application of back-end removal module is shown in Fig.\ref{fig_Back-end}(c).

\section{EXPERIMENTS}

\subsection{Experimental Setup}
To evaluate the performance of our algorithm, we conduct comprehensive comparisons against existing open-source methods on SemanticKITTI\cite{behley2019semantickitti}. 
And to assess the quality of the static map that has been retained after removal of dynamic points, we adopt the preservation rate (PR) and rejection rate (RR) as evaluation metrics\cite{lim2021erasor}. The evaluation metrics are defined as follows:
\begin{equation}
\vspace{-1.5ex}
   PR= \frac{N^{TN}}{N^{sta}}
\vspace{-0.0ex}
\end{equation}
\begin{equation}
 RR=1-\frac{N^{TP}}{N^{dyn}}
\vspace{-1.0ex}
\end{equation}
where $N^{sta}, N^{dyn}$ represent the nums of total static and dynamic points on the ground truth, and $N^{TN}, N^{TP}$ represent the nums of preserved static and dynamic points of generated map. 
F1 score is also computed to reflect the harmonic mean of precision and recall metrics.

Additionally, we conduct real-world experiments of the proposed RH-Map in campus environments with pedestrians and vehicles. Moreover, we integrate RH-Map into open-source navigation system PUTN\cite{jian2022putn} to demonstrate the effectiveness for motion planning of mobile robots.

\begin{figure}[t]
\vspace{-0.3cm}
\centering
\subfloat
{
\centering
\includegraphics[width=0.24\textwidth]{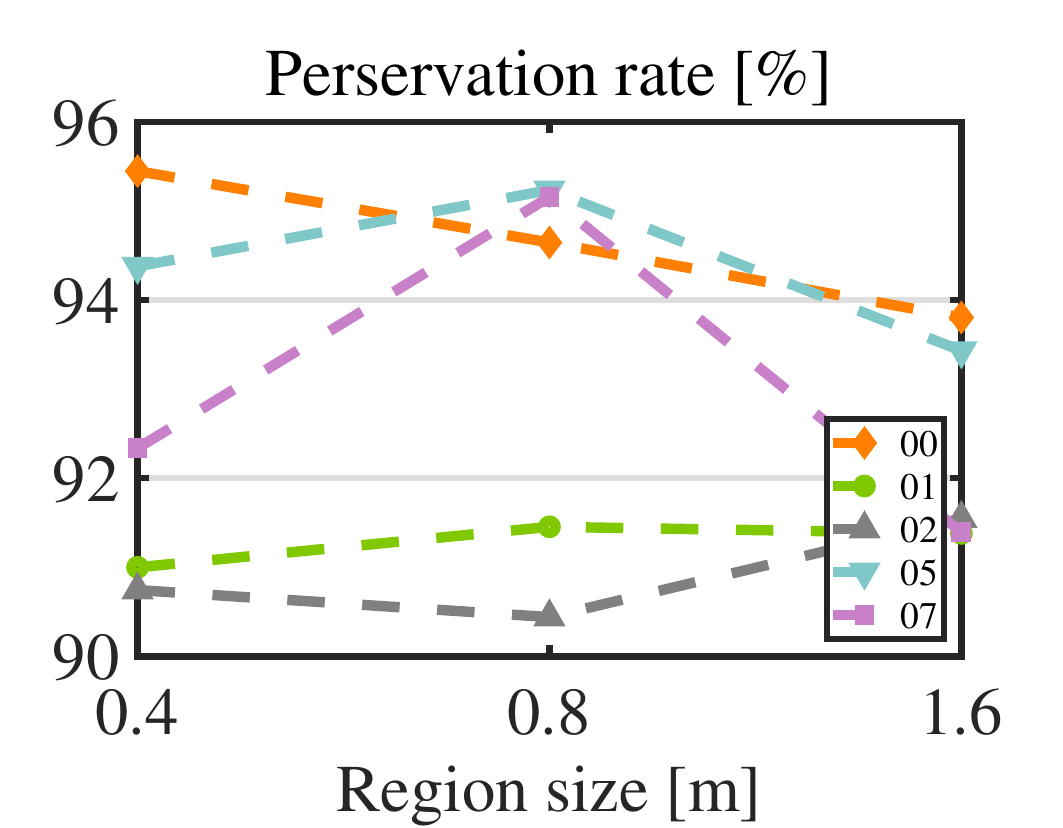}
}
\hspace{-0.6cm}
\subfloat
{
\centering
\includegraphics[width=0.24\textwidth]{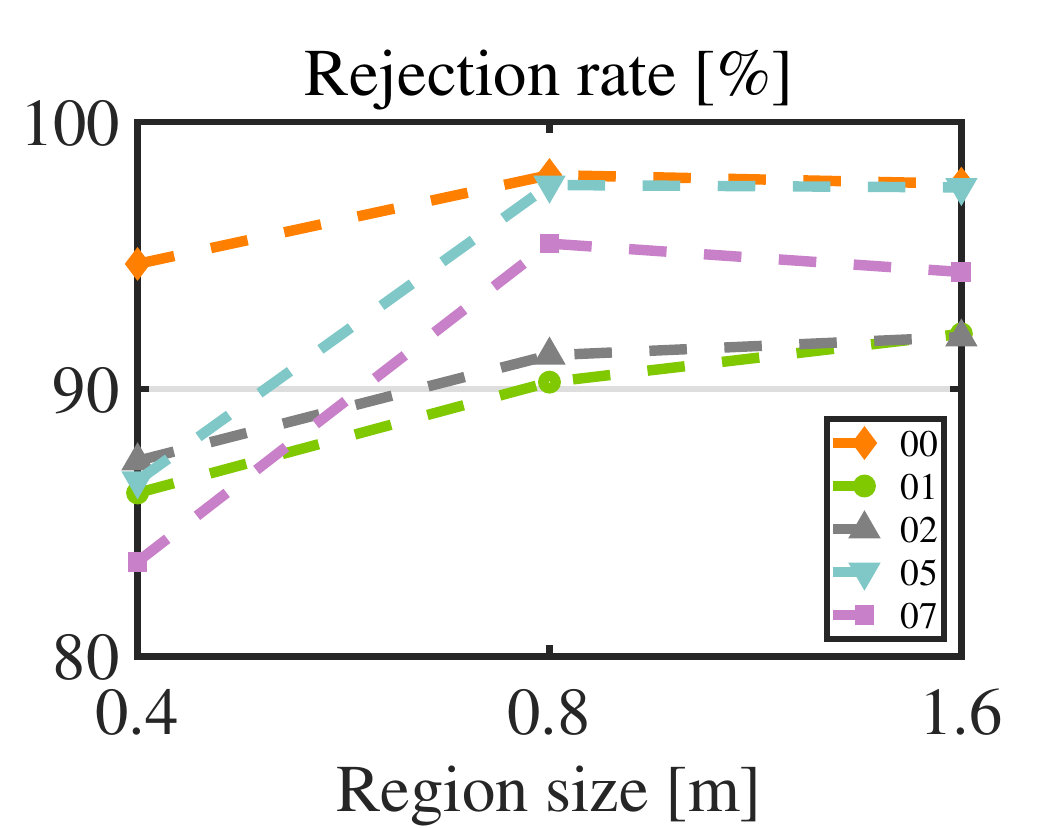}
}
\caption{Performance on SemanticKITTI dataset of different region sizes}
\label{region size}
\vspace{-0.6cm}
\end{figure}
\subsection{Evaluation}
\subsubsection{Impact of region size}
Firstly, we investigate the influence of region size with 0.4 ($mask:2^2-1$), 0.8 ($mask:2^3-1$) and 1.6m ($mask:2^4-1$) on online DOR effect of our RH-map. As shown in Fig.\ref{region size}, it can be observed that increasing region size leads to a decline in PR value, and setting region size as 0.4m  led to a decline in RR value. 
We believe that a smaller region size can lead to inaccurate ground plane fitting, resulting in poor DOR performance. On the other hand, increasing the region size does not significantly improve the DOR performance and actually reduces the retention of static points. We attribute this to the fact that larger regions may misclassify some non-ground static points as dynamic points when removing points except the ground point. As a result, we set 0.8m as the region size for the balance. 

\subsubsection{Comparison with state-of-the-art methods}

\begin{table}[t]   
\vspace{-0.0em}
\begin{center}   
\caption{Comparison with state-of-the-art methods on SemanticKITTI}  
\label{table:1} 
\tabcolsep=3mm
\begin{tabular}{m{0.6cm}<{\centering} m{2.5cm}<{\centering} m{0.7cm}<{\centering} m{0.8cm}<{\centering} m{1.2cm}<{\centering}}   
\hline   \textbf{Seq.} & \textbf{method} & \textbf{PR(\%)} & \textbf{RR(\%)} & \textbf{F1 score} \\   
\hline   \multirow{5}{*}{00}         
&\makecell[l]{OctoMap*\cite{hornung2013octomap} } &57.5321  &\textbf{99.9623}  &0.730317 \\
&\makecell[l]{Removert-RM3*\cite{kim2020remove} }  & 93.2048 &99.3169 &0.961638 \\
&\makecell[l]{ERASOR*\cite{lim2021erasor} }               &  90.5091 &95.9936 &0.978302 \\
&\makecell[l]{RH-Map (Front-end)    }                                     &\textbf{97.8724} &97.788 &\textbf{0.978302} \\
&\makecell[l]{RH-Map (Ours)  }                                       & 94.6446 &98.005 &0.962955 \\
\hline   \multirow{5}{*}{01}         
&\makecell[l]{OctoMap*\cite{hornung2013octomap} } &52.6362 &\textbf{99.8929}  & 0.68944\\
&\makecell[l]{Removert-RM3*\cite{kim2020remove} }&\textbf{95.3201}&95.0313& \textbf{0.951755}\\
&\makecell[l]{ERASOR*\cite{lim2021erasor} }   &89.6322 &94.1192 &0.918209\\
&\makecell[l]{RH-Map (Front-end)}  &92.9481 &77.1859 & 0.843369 \\
&\makecell[l]{RH-Map (Ours)  }&91.454 &90.2589 & 0.908525\\
\hline   \multirow{5}{*}{02}         
&\makecell[l]{OctoMap*\cite{hornung2013octomap} } &41.8728 &\textbf{99.9293} &0.590163 \\
&\makecell[l]{Removert-RM3*\cite{kim2020remove} }  & 79.4783 &93.0953 &0.857496\\
&\makecell[l]{ERASOR*\cite{lim2021erasor} }                 &80.4437&96.2932 &0.876578\\
&\makecell[l]{RH-Map (Front-end)    }                                         &\textbf{98.855} &77.2788 & 0.867 \\
&\makecell[l]{RH-Map (Ours)  }                                                          & 90.4409 &91.2651&\textbf{0.908511} \\
\hline   \multirow{5}{*}{05}         
&\makecell[l]{OctoMap*\cite{hornung2013octomap} }  & 49.1314 & \textbf{99.774} & 0.658409\\
&\makecell[l]{Removert-RM3*\cite{kim2020remove} }  &91.7955 &86.4054 &0.890189 \\
&\makecell[l]{ERASOR*\cite{lim2021erasor} }                & 96.5607 &92.9508 & 0.947214 \\
&\makecell[l]{RH-Map (Front-end)    }                                          &\textbf{97.2321} &96.6255 &\textbf{0.969278} \\
&\makecell[l]{RH-Map (Ours)  }                                                          &94.713 &98.5007 &0.965697 \\
\hline   \multirow{5}{*}{07}         
&\makecell[l]{OctoMap*\cite{hornung2013octomap} } &65.9977 &\textbf{99.2425}  &0.792758 \\
&\makecell[l]{Removert-RM3*\cite{kim2020remove} }& 91.7955 &86.4054 &0.890189\\
&\makecell[l]{ERASOR*\cite{lim2021erasor} }   &91.5798 &  98.536 & 0.949306\\
&\makecell[l]{RH-Map (Front-end)    }   & \textbf{97.3686} &91.2949 & 0.94234\\
&\makecell[l]{RH-Map (Ours)  }& 96.4697 &97.8803 &\textbf{0.971699}\\
\hline  
\bottomrule
\end{tabular}   
   \begin{tablenotes}
     \item[1] "*" means that the algorithm operates offline.
   \end{tablenotes}
\end{center}   
\vspace{-2.5em}
\end{table}
\begin{figure*}[t]
\begin{minipage}[t]{0.19\linewidth}
\centering
\includegraphics[width=3.4cm,height=8cm]{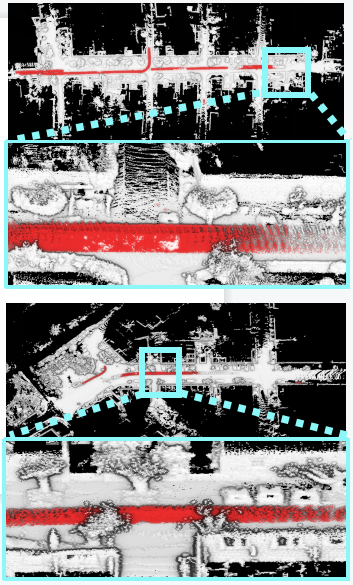}
\centerline{(a) Original map}
\end{minipage}
\begin{minipage}[t]{0.19\linewidth}
{
\centering
\includegraphics[width=3.4cm,height=8cm]{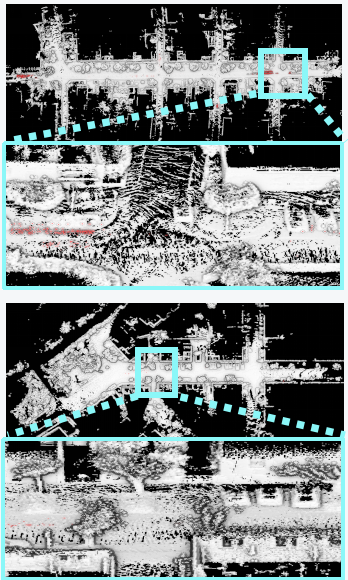}
\centerline{(b) OctoMap\cite{hornung2013octomap}}
}
\end{minipage}
\begin{minipage}[t]{0.19\linewidth}
{
\centering
\includegraphics[width=3.4cm,height=8cm]{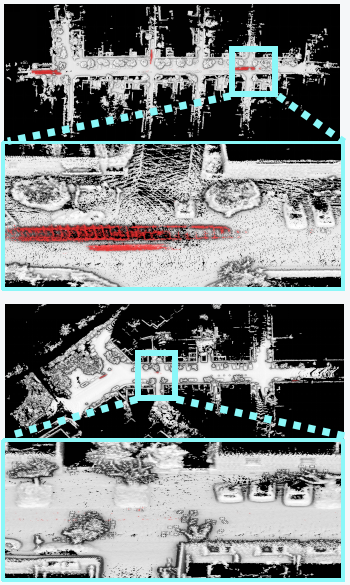}
\centerline{(c) Removert\cite{kim2020remove}}
}
\end{minipage}
\begin{minipage}[t]{0.19\linewidth}
{
\centering
\includegraphics[width=3.4cm,height=8cm]{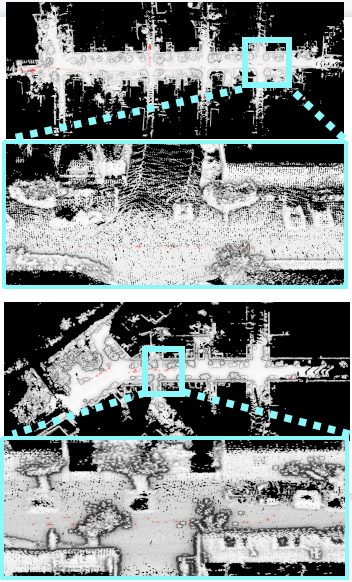}
\centerline{(d) ERASOR\cite{lim2021erasor}}
}
\end{minipage}
\begin{minipage}[t]{0.19\linewidth}
{
\centering
\includegraphics[width=3.4cm,height=8cm]{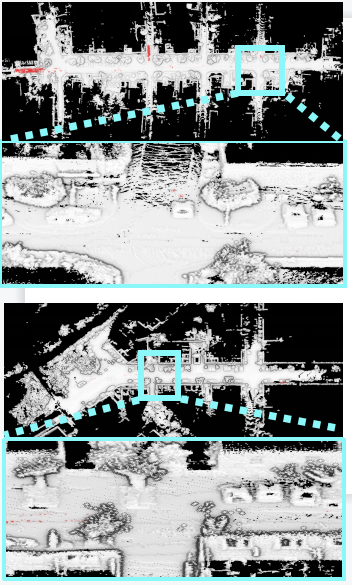}
\centerline{(e) Ours}
}
\end{minipage}
\vspace{-0.3cm}
\caption{Comparison of the final DOR effect conducted by our proposed method and existing methods on sequence 00 and 05 of SemanticKITTI. Red points are dynamic points that are not filtered out. And the fewer red points and more white points preserved, the better effect.}
\label{Kitti comparison}
\vspace{-0.4cm}
\end{figure*}

We conduct the comparative experiments with Octomap\footnote{https://octomap.github.io/}\cite{hornung2013octomap} (voxel size: 0.1m), Removert-RM3\footnote{https://github.com/irapkaist/removert}\cite{kim2020remove} (3 remove stages with pre-built map of voxel size 0.05m), ERASOR\footnote{https://github.com/LimHyungTae/ERASOR}\cite{lim2021erasor} (pre-built map of voxel size 0.1m) and Ours (voxel size: 0.1m). Table \ref{table:1} presents quantitative comparisons of the results of DOR and the quality of  static map can be seen in Fig.\ref{Kitti comparison}. All the results can be reached at the open-source repository.

We perform Octomap offline to achieve its best performance in this experiment due to its poor real-time performance. As described in Table\ref{table:1}, Octomap achieves the highest performance in terms of DOR. However, due to the angle ambiguity of ray-tracing, vast quantities of static points are falsely classified as dynamic, resulting in lower PR values. Removert achieves good results on most sequences, but it generates a significant number of FPs on sequence 02 with slopes, resulting in a relatively low PR value. We believe that it is due to the incidence angle ambiguity of visibility method, which requires more handling of the incident angles.
ERASOR, belonging to segmentation methods, achieves a low PR value on certain sequences due to the use of 2D R-POD descriptors with information loss, leading to some FPs. In contrast, our method is based on a 3D spatial data structure, which helps to alleviate this issue. Note that in the 02 sequence on the highway, there are guardrails in the middle, making it challenging to detect vehicles traveling in the opposite lane, which may hinder our algorithm from achieving better performance. 
Nonetheless, the presence of dynamic objects in the opposite lane has minimal impact on robot navigation.
Obviously, our method achieves comparable performance (ranked first on sequences 00, 02, 05, 07 in F1 score) to that of state-of-the-art approaches (ERASOR, Removert). In addition, our method is free from pre-built map and runs online directly. 
\subsubsection{Ablation study}
We also conduct ablation experiments to test the effectiveness of the back-end removal module, including RH-Map (front-end) and combined front-end and back-end approaches RH-Map(Ours). From Table \ref{table:2}, it can be observed that the front-end-only approach achieves the highest F1 score on sequences 00 and 05. However, in scenes with a higher number of vehicles such as 01, there is a great amount of residual dynamic objects due to the reasons mentioned in \ref{Back-end Removal}. After introducing the back-end module, it can be seen that the RR increases, although it may result in a slight decrease in the PR. Nevertheless, the F1 score remains relatively high and more stable.
\subsubsection{Algorithm speed}
As our algorithm is designed for real-time online processing, we measure the average time and frequency of our proposed algorithm on SemanticKITTI with an Intel(R) Core(TM)i7-12700H CPU only. From Table \ref{table:2}, we can observe that our algorithm achieves average processing frequencies above 10Hz on each sequence, surpassing the typical data publishing rates of LiDAR sensors.

\subsection{Real-world experiments}
\begin{figure*}[t]
    \center
    \includegraphics[width=18cm]{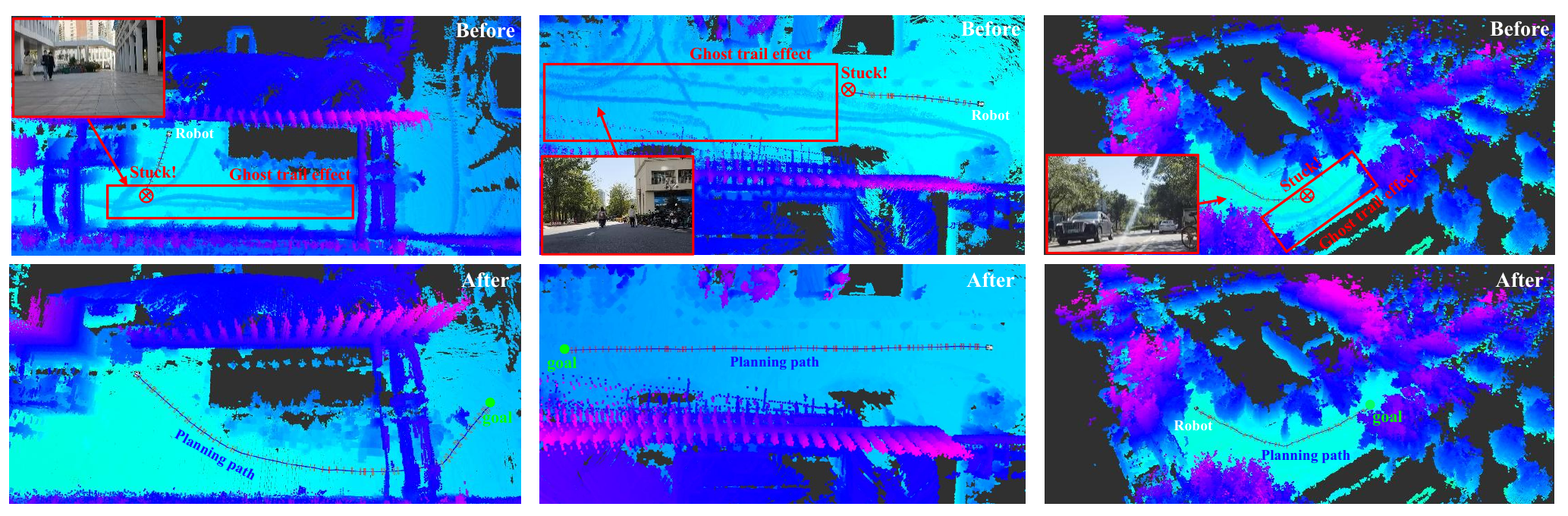}
    \caption{Real-world experiments in different scenes with moving pedestrians and vehicles. The blue curves represent the generated global paths by PUTN.
}
    \label{real-exp}
    \vspace{-0.68cm}
\end{figure*}
To assess the effectiveness and practicality of our method, we conduct real-world experiments in campus environments with pedestrians and vehicles using Scout2 robot base equipped with a 32-beams RS-Helios LiDAR and imu. All the calculation is performed on an Intel(R) Core(TM)i7-12700H CPU which can reach 30 Hz for online DOR with a resolution of 0.1m.
Fast-LIO\cite{xu2022fast} is used to provide $SE(3)$ poses for RH-Map. Fig.\ref{real-exp} displays three scenes with dynamic objects: the top images represent the raw map without DOR, where the red boxes indicate the presence of \textit{ghost trail effect}, and the bottom images demonstrate the online DOR results achieved by our approach.
In addition, we integrate RH-Map with PUTN\cite{jian2022putn} as shown in Fig.\ref{real-exp}: the blue curves represent the generated global paths, and the green dots represent local goals. It can be observed that without DOR on the original map, the navigation algorithm results in sub-optimal global paths or planning failures as shown in the top images. After integrating RH-Map, the robot can navigate successfully and plan effectively in dynamic environments, which is advantageous for outdoor navigation and exploration tasks of mobile robots. More details can be available at \footnote{Video: \url{https://youtu.be/J88xo2M3X6A}.}.
\begin{table}[tb]   
\vspace{-0.7em}
\begin{center}   
\caption{Real-time performance of RH-Map on SemanticKITTI}  
\label{table:2} 
\tabcolsep=3mm
\begin{tabular}{m{0.4cm}<{\centering} m{1.8cm}<{\centering} m{2cm}<{\centering}}   
\hline   
\textbf{Seq.} & \textbf{Runtime (ms)} & \textbf{Frequency (hz)} \\   
\hline   
00 &63.9573 &15.6354 \\
01 &93.2319 &10.7259 \\
02 &66.9781 &14.9303 \\
05 &64.6114 &15.4771 \\
07 &51.0236 &19.5988 \\
\hline  
\bottomrule
\end{tabular}   
\end{center}   
\vspace{-2.7em}
\end{table}

\section{CONCLUSION}
In this paper, a novel online map construction framework of DOR in dynamic environments, RH-Map, is proposed. Based on the proposed region-wise hash map data structure, RH-Map consists of a real-time front-end module Scan Fresher for ground estimation and online DOR, and a lightweight back-end module for further removal of residual dynamic objects. Comparative validations on SemanticKITTI and real-world experiments are conducted to evaluate the proposed method. The experimental results demonstrate the efficiency of RH-Map in online map construction of DOR.

\bibliographystyle{IEEEtran}
\renewcommand{\baselinestretch}{0.887}
\bibliography{IEEEabrv,bib/bibliography}
\end{document}